# Character-level Chinese Writer Identification using Path Signature Feature, DropStroke and Deep CNN

Weixin Yang, Lianwen Jin[+], Manfei Liu
College of Electronic and Information Engineering
South China University of Technology
Guangzhou, China
wxy1290@163.com, +lianwen.jin@gmail.com, liu.manfei@mail.scut.edu.cn

*Abstract*—Most existing online writer-identification systems require that the text content is supplied in advance and rely on separately designed features and classifiers. The identifications are based on lines of text, entire paragraphs, or entire documents; however, these materials are not always available. In this paper, we introduce a path-signature feature to an end-to-end text-independent writer-identification system with a deep convolutional neural network (DCNN). Because deep models require a considerable amount of data to achieve good performance, we propose a data-augmentation method named DropStroke to enrich personal handwriting. Experiments were conducted on online handwritten Chinese characters from the CASIA-OLHWDB1.0 dataset, which consists of 3,866 classes from 420 writers. For each writer, we only used 200 samples for training and the remaining 3,666. The results reveal that the path-signature feature is useful for writer identification, and the proposed DropStroke technique enhances the generalization and significantly improves performance.

*Keywords—Deep convolutional neural network; online text-independent writer identification; data augmentation; path-signature feature; Chinese character*

## I. Introduction

Owing to the development and commercialization of pen-based interfaces and touch-screen mobile terminals, online writer-identification systems are getting increased attention for personal identification, digital forensics, and downstream applications. Writer-identification systems are widely applicable in identifying individuals, authenticating access to a network, performing mobile bank transactions, distinguishing forensic trace evidence and so on [2]. It is often desirable for these systems to be exclusively responsible for security, and for the identification process to be automatic and immediate. To fulfill these requirements, numerous advanced writer-identification methods have been proposed, investigated, and developed, yielding high-level accuracy [2-7].

Writer identification generally follows the pipeline of data acquisition, pre-processing, feature extraction, and feature matching or classification [3]. Depending on how the handwritings are inputted, the identification process can be either offline and online. Online methods are considered less challenging because of the availability of rich information such as pen-position, pen-down or pen-up, pen-velocity, pen-pressure, pen-altitude, and pen-azimuth. Offline methods, by contrast, merely have access to scanned images [4][5]. Features are generally extracted in two ways: text-dependent or text-independent. Text-dependent systems are highly accurate, but they are inapplicable in cases where the text content is absent. In comparison, text-independent systems have a wider range of application, but they often require a large amount of data to ensure a sufficient generalization capacity to deal with natural handwriting [6]. To evaluate handwriting, different levels of material in different languages are used. Materials consisting of a single character, a line of text (10~20 characters), a paragraph (50~100), or an entire document (100 or more) lead to varying difficulties in terms of acquiring the information needed for identification. Among the multilingual materials used for identification, Chinese handwriting is regarded as the most difficult, owing to the enormous number of characters (over 50,000) and the complex stroke structures [7]. Despite the tremendous research on writer identification, there is relatively little research available in online text-independent Chinese character-based handwriting.

In recent years, deep convolutional neural networks (DCNNs) have been considerably influential in overcoming many challenges related to computer vision and pattern recognition, such as image detection and classification [8][9], face verification [10][11], text spotting [12][13], and handwriting recognition [14][15]. DCNNs consistently outperform traditional methods by a large margin. In the field of Chinese handwriting recognition, Graham [15] proposed a variation of the CNN called DeepCNet, which takes advantage of spatial sparsity and slow convolutional and max-pooling layers rather than fast layers. By considering the handwriting trajectory as a finite path and then adopting a path-signature feature [16] as the representation, DeepCNet achieved breakthrough results and won first place at the 2013 International Conference on Document Analysis and Recognition in the online handwritten Chinese character recognition competition [17]. However, to the best of our knowledge, although writer identification is similar to handwriting recognition, little research has been reported on identification employing deep neural networks.

Online Chinese handwriting identification remains a challenge because of several factors. First, it is inconvenient for applications to collect an entire line of text, paragraphs, or complete documents in order to ensure confidence in the results. Second, the vocabulary of the text is restricted by the insufficiency of the generalization capability. Even though text-independent methods are entirely content-free, their ability to represent and discriminate is limited with unconstrained handwriting. Third, each step in the identification process is separately designed, optimized, and performed, resulting in an unbalanced pipeline with no knowledge of the deficient parts. Fourth, making identification decisions is increasingly difficult with materials consisting of single characters or even scrawling strokes that are ostensibly devoid of valuable information.

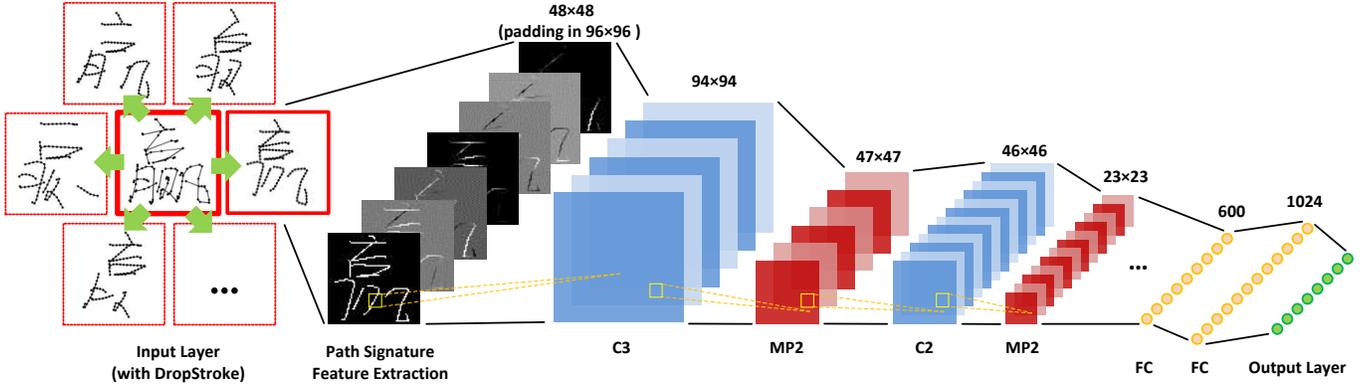

Fig. 1. Illustration of the DCNN for writer identification enhanced by DropStroke and a path-signature feature.

To alleviate these problems, we introduce a path-signature feature to writer identification because of its ability to uniquely characterize the trajectory of a character. The extracted feature maps are followed by a DCCN—an end-to-end system enabling integrated training that provides overall optimization. Moreover, we propose a new technique, namely DropStroke, which randomly omits some strokes in the characters. This method can augment the data and enhance the generalization capability of the network greatly. The schematic for our system is given in Fig. 1.

The remainder of this paper is organized as follows. Section II presents the architecture of a DCNN. Section III describes the feature for extracting the path signature. The DropStroke method is presented in Section IV. Section V presents the experimental results and analysis. Finally, conclusions are summarized in Section VI.

## II. DEEP CONVOLUTIONAL NEURAL NETWORK

Graham proposed a DCNN architecture called DeepCNet [15] for high performance recognition of images. DeepCNet takes advantage of a sparseness of input samples and uses slow convolutional filters for the reason that a slow convolutional speed can retain more spatial information and enhance generalization. Our DCNN inherits these advantages and has an architecture similar to [15], but includes an extra Fully-Connected (FC) layer. As shown in Figure 1, the structure of our DCNN consists of six convolutional layers, the first five of which are followed by Max-Pooling (MP) layers. The size of the convolutional filter is $3\times3$ for the first layer and $2\times2$ for the others, with a constant stride of 1 pixel. Max-pooling is carried out over a $2\times2$ pixel window with a stride of 2 pixels. At the top of the network, two FC layers, respectively 600 and 1024 pixels in size, are designed to characterize the complex structure of Chinese characters and provide more nonlinearity to the network as compared with DeepCNet. The number of convolutional filter kernels is set to 100 for the first layer and increases by an increment of 100 after each max-pooling. Rectified linear units (ReLUs) [18] are used as activation functions for neurons in all convolutional layers and FC layers, whereas the softmax is used for the output layer.

We render the image of the input character as a $48\times48$ bitmap image embedded in a $96\times96$ grid that is initialized to zero for spatial padding. Thus, the architecture for our DCNN can be represented as Input-$M\times96\times96$-80C3-MP2-160C2-MP2-240C2-MP2-320C2-MP2-400C2-MP2-480C2-512FC-Output, where $M$ denotes the number of input channels, which varies from 1 to 15 depending on the signature features discussed in Section III.

## III. PATH-SIGNATURE FEATURE EXTRACTION

The path-signature feature, pioneered by Chen [16] in the form of iterated integrals, can extract a sufficient quantity of information contained in a finite path (e.g., in online handwriting) to solve any linear differential equation. The path-signature feature was first introduced for the recognition of online handwritten characters by Graham [15]. In this paper, we discovered a method for extracting sufficient writer information from handwriting using the path-signature feature.

Assume a time interval $[T_1, T_2] \subset \mathbb{R}$ and the writing plane $W = \mathbb{R}^2$. A pen stroke can be expressed by a continuous function $P:[T_1,T_2] \to W$. Given the intervals $[t_1,t_2] \subset [T_1,T_2]$ and $k \in \mathbb{N}^*$, the $k$-th iterated integral of $P$ is the $2^k$ dimensional vector defined by:

$$P^k_{t_1,t_2} = \int_{t_1<u_1<\cdots<u_k<t_2} 1 dP_{u_1} \otimes \cdots \otimes dP_{u_k}. \qquad (1)$$

Conventionally, the $k=0$ iterated integral denotes the original input, and $k=1$ represents path displacement, whereas $k=2$ denotes path curvature. By increasing $k$, higher levels of path information can be extracted, but the dimensions of iterated integrals enlarge rapidly as well. Therefore, truncated signatures are often used in practice. The signatures

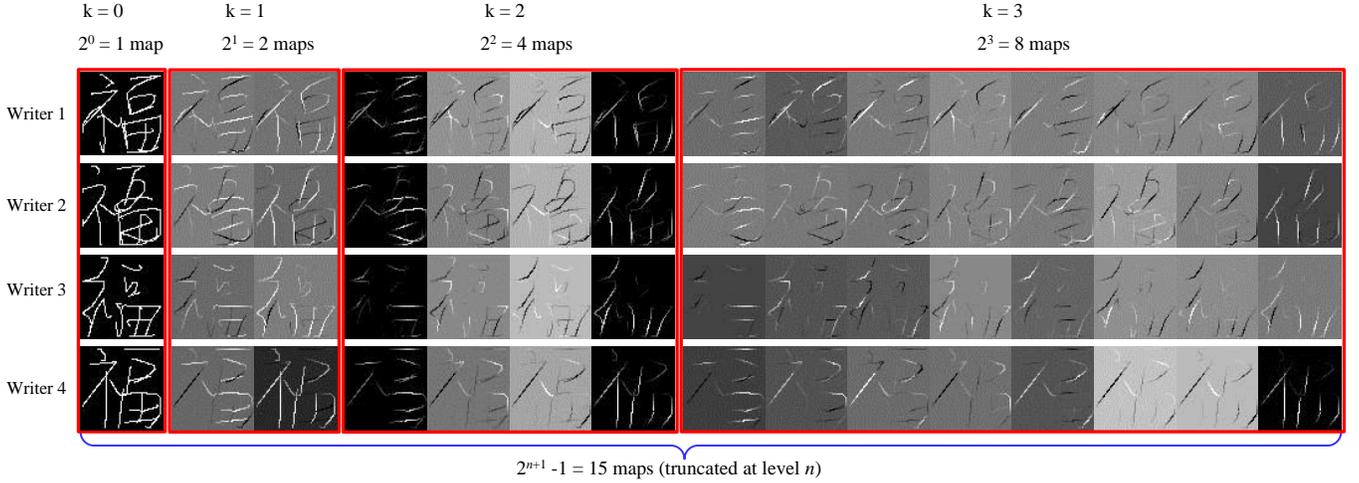

Fig. 2. Visualization of the path-signature feature with a truncated level $n$.

truncated at level $n$ are the combination of the iterated integrals as:

$$S(P)_{t_1,t_2}^n = (1, P_{t_1,t_2}^1, P_{t_1,t_2}^2, \ldots, P_{t_1,t_2}^n). \quad (2)$$

The iterated integrals $P_{t_1,t_2}^k$ in (1) can be calculated by:

$$P_{t_1,t_2}^0 = 1,\ P_{t_1,t_2}^1 = \Delta_{t_1,t_2},\ P_{t_1,t_2}^2 = (\Delta_{t_1,t_2} \otimes \Delta_{t_1,t_2})/2!,$$
$$P_{t_1,t_2}^3 = (\Delta_{t_1,t_2} \otimes \Delta_{t_1,t_2} \otimes \Delta_{t_1,t_2})/3!,\ \ldots, \quad (3)$$

where $\Delta_{t_1,t_2} := P_{t_2} - P_{t_1}$ denotes the path displacement. At this stage, each point along the path can generate a set of signature values with the truncated level $n$. The dimension of $S(P)$ is calculated by $2^{n+1}-1$ (i.e., the number of feature maps).

For better observation, we fill the pen's trajectory with the respective signature values at each level $k$ from 0 to 3, and set the background pixels to 0. We then apply the image histogram equalization for each feature map, the visualization for which is shown in Fig. 2. Each row corresponds to the handwriting feature maps contributed by a writer. Each column can be regarded as a specific descriptor that extracts special information (e.g., the direction or curvature) from the original path. For implementation (see Section V), we evaluated the first four (from the zeroth) iterated integrals signature features to compare their effects.

## IV. DropStroke for Writer Identification

Text-independent identification systems and deep learning models both require sufficient training data to ensure a generalization capability, but collecting an extensive amount of personal handwriting is inconvenient and likely vexing to users in practical applications. Therefore, inspired by Dropout [19], which randomly omits part of the feature detectors in each training case, we propose a data-augmentation method called DropStroke for writer identification. By randomly dropping some strokes from an original character, a set of new handwritten characters is generated with the combination of remaining strokes.

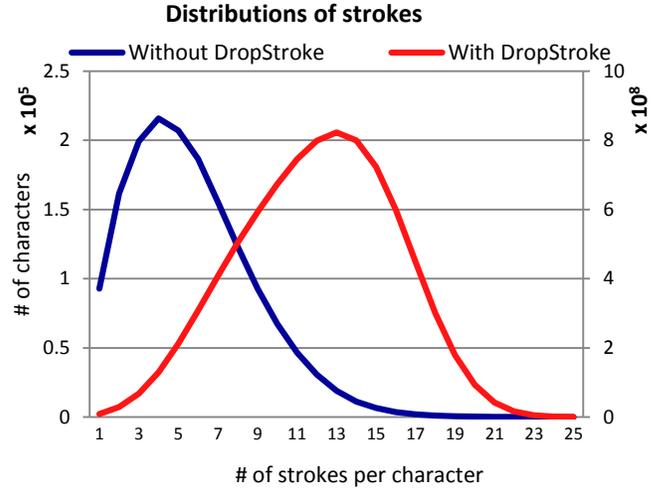

Fig. 3. Distribution of strokes with and without DropStroke on CASIA-OLHWDB1.0 [1]. The y-axis on the left denotes the # of characters before DropStroke, and the y-axis on the right after applying DropStroke.

There are several advantages to DropStroke. Suppose that $m$ strokes are dropped out from an $n$-stroke handwritten character, the number of new possible characters is:

$$C_n^m = n!/m!(n-m)!. \quad (4)$$

Thus, the number of all the generated characters is the sum of (4) with all $m$ $(0 \leq m < n, m \in \mathbb{Z}^+)$, such that

$$\sum_{m=0}^{n-1} C_n^m = (\sum_{m=0}^n C_n^m) - C_n^n = 2^n - 1. \quad (5)$$

Using (4) and (5) with all the characters in the CASIA-OLHWDB1.0 dataset [1], Fig. 3 illustrates the effect on the distribution of strokes. With DropStroke, the number of generated characters increases more than 5,000-fold, from $1.6 \times 10^6$ to $8.5 \times 10^9$, demonstrating considerable augmentation. In addition, the DropStroke method breaks the original stroke structure and prevents structural information from being considered; these personal stroke structures may to some extent lead to better identification, but they limit the generalization faced by handwriting with diverse stroke structures (e.g., in natural or multilingual handwritings), and therefore DropStroke can enhance the generalization. Moreover, the same test sample can generate new characters with DropStroke and can be identified multiple times to obtain a number of different predictions, so that averaging these predictions can further improve accuracy.

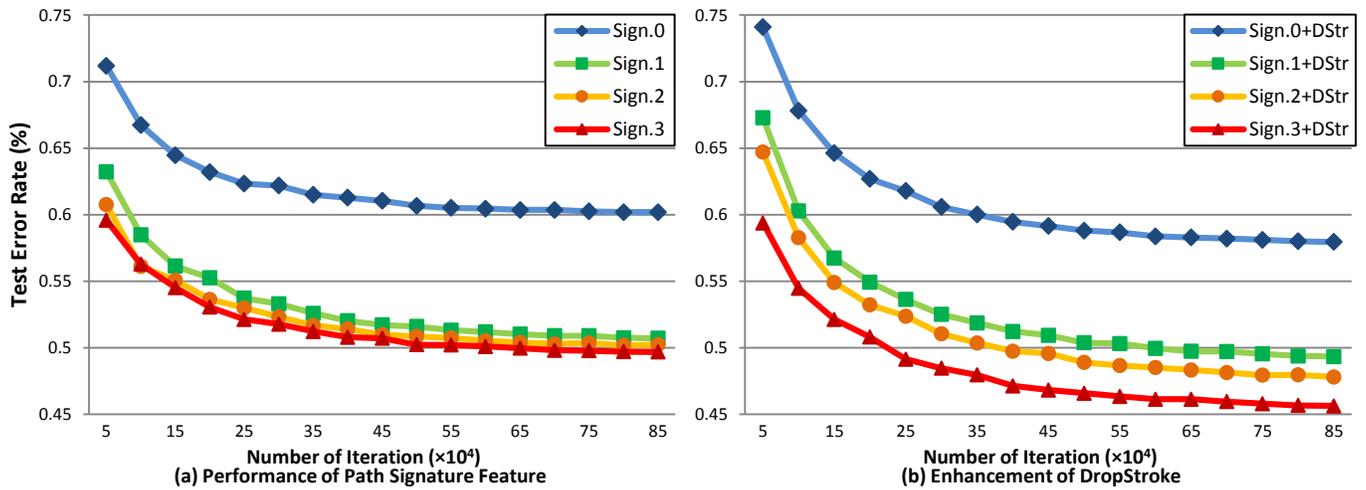

Fig. 4. Performance of different truncated versions of path signature feature and their enhancements with DropStroke. The results are the Top-1 error rates evaluated on Chinese character-level handwritings on CASIA-OLHWDB1.0 [1] and the test samples use the prototypes without DropStroke.

However, as indicated by (5), an obvious disadvantage to DropStroke is that the number of generated characters only depends on the number of strokes $n$ from the prototypes. Owing to the diversity of writing habits, different writers contribute varying numbers of strokes for the same character, accounting for the uneven training data of each writer after applying DropStroke. To alleviate this, the training samples are randomly selected from the prototypes before DropStroke is applied to each of them.

## V. EXPERIMENTAL RESULTS AND ANALYSIS

The most popular online Chinese handwriting identification database is the National Laboratory of Pattern Recognition (NLPR) handwriting database [20] in which the samples are provided in pages and contain information such as position, pen-down and pen-up state, azimuth, altitude and pressure. However, in order to evaluate our system at the level of an individual character, and to simulate the general conditions when only position and time information are available, we used the online handwritten Chinese character dataset CASIA-OLHWDB1.0 [1], consisting of 3,866 classes contributed by 420 writers using the same writing device. We randomly selected 200 classes for training, and the remaining 3,666 were used for testing to fulfill text-independence.

In our experiments, we used the single DCNN structure shown in Section II. A random mix of affine transformations (i.e., translation, rotation, and scaling) was adopted as the elastic distortion operation for data augmentation during the training stage. The training mini-batch size was set to 100, and the Dropout [19] rate for the last four weighting layers was experimentally determined at 0.1, 0.1, 0.5, and 0.5, respectively. We performed our experiments on a PC with a GTX780 GPU and spent a week training the DCNN system.

TABLE I. IDENTIFICATION ACCURACY RATES (%) WITH VARIOUS METHODS

| Rank | Methods | | |
|---|---|---|---|
| | *Sign.3* (*1 test*) | *Sign.3 + DStr* (*1 test*) | *Sign.3 + DStr* (*20 test*) |
| *Top-1* | 50.19 | 54.06 | **55.45** |
| *Top-5* | 73.65 | 76.85 | **77.80** |
| *Top-10* | 81.62 | 84.17 | **84.91** |
| *Top-15* | 85.63 | 87.79 | **88.36** |
| *Top-20* | 88.13 | 90.03 | **90.46** |

Primarily, we conducted experiments to compare the path-signature features (denoted by Sign.$n$ at truncated version $n$) in different truncated versions, as shown in Fig.4(a). The feature map for Sign.0 is exactly an offline bitmap, as seen in the first column of Fig. 2. Even though the Sign.0 contains only position information, the DCNN performed well, indicating that the network itself has a satisfactory generalization. With the availability of temporal information, Sign.1, Sign.2, and Sign.3 significantly improved the network, reaching an accuracy rate of 50%. This demonstrates that the path-signature feature is very useful for distinguishing different writers.

Moreover, we evaluated the proposed DropStroke (abbreviated to DStr) with each of the foregoing experiments in Fig.4(b). The curves of the error rates appear high at the beginning because the size of training set is enlarged and the generated samples are less predictable. These curves converge quickly, and ultimately achieve excellent results. With the use of DropStroke, Sign.3 actually performed better than the others, whereas this difference is negligible in the case without DropStroke. Our approach is flexible insofar as it randomly generates multiple test samples from the prototypes to offer several predictions. We averaged 20 of these predictions and further compared the identification accuracy with different methods, the results for which are listed in Table I.

TABLE II. IDENTIFICATION ACCURACY RATES (%) OF MIMIC MATERIALS ON VARIOUS LEVELS

| Rank | Mimic Materials on Various Levels | | | | | | | | |
|---|---|---|---|---|---|---|---|---|---|
| | Character | Chinese Word (2~3 characters) | | Chinese Phrase (4~6 characters) | | | Chinese Text Line (10~20 characters) | | |
| | 1 | 2 | 3 | 4 | 5 | 6 | 10 | 15 | 20 |
| Top-1 | 55.45 | 71.27 | 80.33 | 86.27 | 90.10 | 92.77 | 97.58 | 99.07 | 99.52 |
| Top-5 | 77.80 | 90.28 | 94.93 | 97.10 | 98.17 | 98.82 | 99.71 | 99.93 | 99.98 |
| Top-10 | 84.91 | 94.62 | 97.64 | 98.82 | 99.33 | 99.61 | 99.92 | 99.98 | 100.00 |
| Top-15 | 88.36 | 96.37 | 98.56 | 99.33 | 99.62 | 99.79 | 99.97 | 99.99 | 100.00 |
| Top-20 | 90.46 | 97.27 | 98.99 | 99.55 | 99.75 | 99.85 | 99.98 | 99.99 | 100.00 |

Furthermore, character-level identification systems are also available for evaluating materials at other levels, such as Chinese words (2~3 characters), phrases (4~6), lines (10~20), though combining the predictions of the characters. Hence, we sequentially divided the training set into groups, each of which contained the same number of characters, to form the mimic materials at a certain level. Using our above-mentioned approach (Sign.3+DStr (20 test)), the results for the mimic materials on various levels are given in Table II. Previous best approach achieved an accuracy rate of 92% [5], the highest reported using the NLPR database [20], and this was achieved by testing on pages (50 characters per page) of text from 242 writers using rich online information such as pressure, altitude and azimuth besides position and time information. However, our approach achieved an even higher accuracy rate of 99.52% on mimic text lines (20 characters per line) from 420 writers with access only to position and time information using the CASIA-OLHWDB dataset.

VI. CONCLUSION

This paper presented a deep CNN system to address the online Chinese character-level writer-identification problem. Incorporated with a path-signature feature and the proposed DropStroke, the performance improved significantly. This indicates that the path-signature feature is useful for distinguishing handwriting, and that DropStroke greatly enlarges the size of the dataset to enhance the generalization of our system. The results of the mimic materials on a line of text achieved an accuracy rate of 99.52%.

Although the DropStroke method proposed is designed for writer identification, we believe that it can serve as a general data-augmentation technique that could be extended to other tasks (e.g., for character recognition, gender identification). Such potential applications merit further investigation.

ACKNOWLEDGMENT

This research is supported in part by NSFC (Grant No.: 61472144), National science and technology support plan (Grant No.:2013BAH65F01 -2013BAH65F04), GDSTP (Grant No.: 2012A010701001, 2012B091100396), GDUPS(2011), Research Fund for the Doctoral Program of Higher Education of China (Grant No.: 20120172110023).

REFERENCES

[1] C.L. Liu, F. Yin, D.H. Wang and Q.F. Wang, "CASIA online and offline Chinese handwriting databases," Proc. 11th Int'l Conf. Document Analysis and Recognition, pp. 37-41, 2011.

[2] A. Chaabouni, H. Boubaker, M. Kherallah, H. El-Abed and A. Alimi, "Static and dynamic features for writer identification based on multi-fractals," The International Arab Journal of Information Technology, 2013.

[3] H.E. Said, T.N. Tan and K.D. Baker, "Personal identification based on handwriting," Pattern Recognition, vol. 33, no 1, pp. 149-160, 2000.

[4] S.M. Idicula, "A survey on writer identification schemes," Writer, vol. 25, no. 12, pp. 15, 2011.

[5] B. Li, Z. Sun and T. Tan, "Hierarchical shape primitive features for online text-independent writer identification," Proc. 10th Int'l Conference on Document Analysis and Recognition, pp. 986-990,2009.

[6] B. Li and T.N. Tan, "Online text-independent writer identification based on temporal sequence and shape codes," Proc. 10th Int'l Conference on Document Analysis and Recognition, pp.931-935, 2009.

[7] W.Y. Leng and S.M. Shamsuddin, "Writer identification for Chinese handwriting," Int. J. Advance. Soft Comput. Appl, 2(2): 142-173, 2010.

[8] A. Krizhevsky, I. Sutskever and G.E. Hinton. "Imagenet classification with deep convolutional neural networks," Advances in neural information processing systems, pp. 1097-1105, 2012.

[9] O. Russakovsky, J. Deng, H. Su, et al., "Imagenet large scale visual recognition challenge," arXiv preprint arXiv:1409.0575, 2014.

[10] Y. Taigman, M. Yang, etal , "Deepface: Closing the gap to human-level performance in face verification," IEEE CVPR, pp. 1701-1708, 2014.

[11] Y. Sun, Y. Chen, X. Wang and X. Tang. "Deep learning face representation by joint identification-verification," Advances in neural information processing systems, pp. 1988-1996, 2014.

[12] A. Coates, B. Carpenter, C. Case, et al., "Text detection and character recognition in scene images with unsupervised feature learning," Proc. 11th Int'l Conf. Document Analysis and Recognition, pp. 440-445, 2011.

[13] M. Jaderberg, K. Simonyan, A. Vedaldi and A. Zisserman, "Reading text in the wild with convolutional neural networks," arXiv preprint arXiv:1412.1842, 2014.

[14] D.C. Cireșan, U. Meier and J. Schmidhuber, "Multi-column deepneural networks for image classification," IEEE Conf. Computer Vision and Pattern Recognition, pp. 3642-3649, 2012.

[15] B. Graham, "Sparse arrays of signatures for online character recognition," arXiv preprint arXiv:1308.0371, 2013.

[16] K.T. Chen, "Integration of paths--A faithful representation of paths by noncommutative formal power series," Trans. of the American Mathematical Society, pp. 395-407, 1958.

[17] F.Yin, Q.F. Wang, X.Y. Zhang and C.L. Liu, "ICDAR 2013 chinese handwriting recognition competition," Proc. ICDAR, pp. 1464-1470, 2013.

[18] V. Nair and G.E. Hinton, "Rectified linear units improve restricted boltzmann machines," Proc. 27th ICML, pp. 807-814, 2010.

[19] G.E. Hinton, N. Srivastava, A. Krizhevsky, I. Sutskever and R.R. Salakhutdinov, "Improving neural networks by preventing co-adaptation of feature detectors," arXiv preprint arXiv: 1207.0580, 2012.

[20] NLPR datadbase. http://www.cbsr.ia.ac.cn/english/Databases.asp.